# Three-Dimensional Investigation of the Metric Properties of Parabolic Double Projection Involving Catadioptric Camera

Ahmed Hamdy[1], Ahmed Elsherif[2], Saiid Shebl[3]

Alex. Univ., Faculty of Eng., Egypt

Email: ahmed_hamdy@alexu.edu.eg, ahmed.elsherif@alexu.edu.eg, sshebl@hotmail.com,

*Abstract*: This paper presents an analytical study for the metric properties of the paraboloidal double projection, i.e. central and orthogonal projections used in the catadioptric camera's system. Metric properties have not sufficiently studied in previous treatments of such system. These properties incorporate the determination of the true lengths of projected lines and areas bounded by projected lines. The advantageous main gain of determining metric elements of the paraboloidal double projection is studying distortion analysis and camera calibration, which is considered an essential tool in testing camera accuracy. Also, this may be considered as a significant utility in studying comparison analysis between different cameras projection systems.

## 1) Introduction

The parabolic double projection has been previously studied by many authors both graphically and analytically [1]. Such projection is carried out centrally on the parabolic surface as projection surface from the surface focus as the center of the projection, then orthogonally on the directory plane $\tau$.

This double projection may be considered as the most relevant projection system for the catadioptric camera imaging since it provides wide viewing coverage [2,3]. Although the necessity of metric properties determination of the projection for the purposes of camera calibration and accuracy assessment [4], to the authors' opinion, metric properties of such projection has never been focused on previous treatments. This paper originates genuine analytical study for the determination of the metric properties of the parabolic double projection. Such properties include the determination of the true lengths of projected lines and area bounded by projected lines on both the surface and on the directory plane.

## 2) The Concept of Parabolic Double Projection

Figure.1 exhibits the technique tracked for establishing the perspective double projection of point *A* onto the paraboloid surface and the directory plane $\tau$. In the figure, $\sigma$ is the focal plane, the plane of the *u-v* axes which passes through surface focus *O*. The plane $\tau$ is the directory plane which is parallel to $\sigma$ and the distance between $\tau$ and $\sigma$ equals *2f*. where, *2 f* is the surface focal length. Point *A* is a space point of coordinates ($U_A$, $V_A$, $W_A$) where the *w*-axis is perpendicular to $\sigma$ and $A^*$ is the orthogonal projection of *A* onto the focal plane $\sigma$. Also, in the Figure, $\theta_A$ is the angle of inclination of *OA* to $\sigma$ and $\Phi_A$ is the angle between $OA^*$ and *u*-axis. let the paraboloid surface Equation be [1]:

$$U^2 + V^2 = 4f \cdot W + 4f^2 \qquad \text{Eq. (1).}$$

Then, The double projection technique in Figure.1. may be carried out as follows [1]:

- *A* ($U_A$, $V_A$, $W_A$) is joined to the focus *O*, then the ray *AO* intersects the surface at the perspective $A_1(U_{A_1}, V_{A_1}, W_{A_1})$, where:

$$U_{A_1} = \frac{2f \cdot U_A (W_A \pm \sqrt{W_A^2 + r_A^2})}{r_A^2} \qquad \text{Eq. (2).}$$

$$V_{A_1} = \frac{2f \cdot V_A (W_A \pm \sqrt{W_A^2 + r_A^2})}{r_A^2} \qquad \text{Eq. (3).}$$

$$W_{A_1} = \frac{2f \cdot W_A (W_A \pm \sqrt{W_A^2 + r_A^2})}{r_A^2} \qquad \text{Eq. (4).}$$

And

$$r_A = \sqrt{U_A^2 + V_A^2} \qquad \text{Eq. (5).}$$

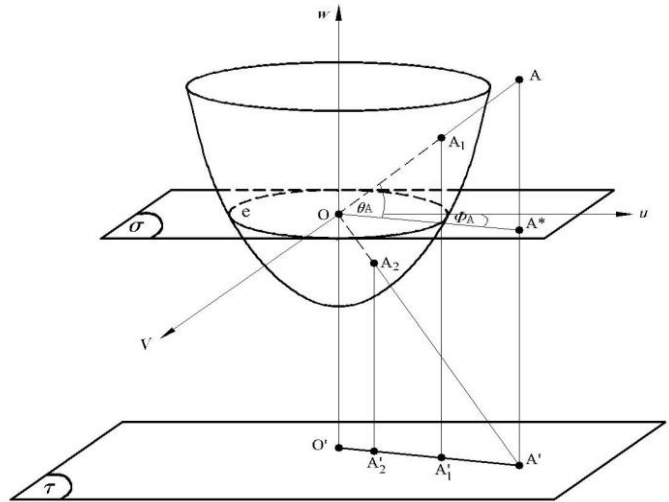

**Figure.1. Paraboloid double projection.**

i.e., the space points are projected centrally from *O,* as center of projection, into the paraboloid as a projection surface.
- *A* is projected orthogonally onto the directory plane $\tau$ to obtain the horizontal projection *A'* ($U_{A'}$, $V_{A'}$, $W_{A'}$) of *A*, where:
$$U_{A'} = U_A, \ V_{A'} = V_A, and \ W_{A'} = -2f \qquad \text{Eq. (6).}$$
- $A_1$ is projected orthogonally onto the directory plane $\tau$ to get the horizontal projection *A'₁* of $A_1$, where:
$$U_{A'_1} = U_{A_1}, V_{A'_1} = V_{A_1} \ and \ W_{A'_1} = -2f \qquad \text{Eq. (7).}$$
- Point *A'* is joined to the focus *O*, then the ray *A'O* intersects the surface at the perspective $A_2$ of *A'*, where:

$$U_{A_2} = \frac{2f \cdot U_A \ (-2f \ \pm \sqrt{4f^2 + r_A^2} \ )}{r_A^2} \qquad \text{Eq. (8).}$$

$$V_{A_2} = \frac{2f \cdot V_A \ (-2f \ \pm \sqrt{4f^2 + r_A^2} \ )}{r_A^2} \qquad \text{Eq. (9).}$$

$$W_{A_2} = \frac{-4f^2 \ (-2f \ \pm \sqrt{4f^2 + r_A^2} \ )}{r_A^2} \qquad \text{Eq. (10).}$$

- $A_2$ is projected orthogonally onto the directory plane $\tau$ to obtain the horizontal projection *A'₂* of $A_2$, where:
$$U_{A'_2} = U_{A_2}, V_{A'_2} = V_{A_2} \ and \ W_{A'_2} = -2f \qquad \text{Eq. (11).}$$

As a conclusion, there exists two consequent central projections for any point *A*, the first is the perspective projection of the point on the paraboloid surface, then, the second is the perspective $A_2$ of the orthogonal projection of *A*.

## *2.1 ) Metric Properties of Parabolic Double Projection*

Metric properties of the parabolic double projection are of great importance in the favor of camera calibration and accuracy assessment. Although camera geometric control is considered an essential tool for camera calibration [5], metric properties of the projection never been handled before. Therefore, the analysis concerns determining metric properties of such projection presented in this paper is considered as genuine analysis for imaging concept. This paper may be originating analytical study for the determination of the metric properties of the parabolic double projection including the determination of the true lengths of projected lines and area bounded by projected lines on both paraboloid surface and on the directory plane.

## *2.1.1 ) The True Length of perspectively and Orthogonally Projected Lines*

In Figure. 2, it is desired to determine the following true lengths:
- **true length of space line *L*** joins two given points *A* and *B* which may be expressed by the equation:
$$L = \sqrt{(U_A - U_B)^2 + (V_A - V_B)^2 + (W_A - W_B)^2} \qquad \text{Eq. (12).}$$
- **true length of the line *L'***, the orthogonal projection of the space line *L* on the directory plane. This may be expressed as:
$$L' = \sqrt{(U_{A'} - U_{B'})^2 + (V_{A'} - V_{B'})^2} \qquad \text{Eq. (13).}$$
- **true length of the perspective projection of a line on the paraboloid surface.**

Figure. 2 represent a general line in space and its projections. Where, *L₁* is the perspective projection of the line

---

$L$ into the paraboloid surface. Such a perspective is the curve of intersection of the surface and the plane $\alpha$ [$O, L$]. Hence, the planar intersection of paraboloid must be analyzed to classify such intersections.

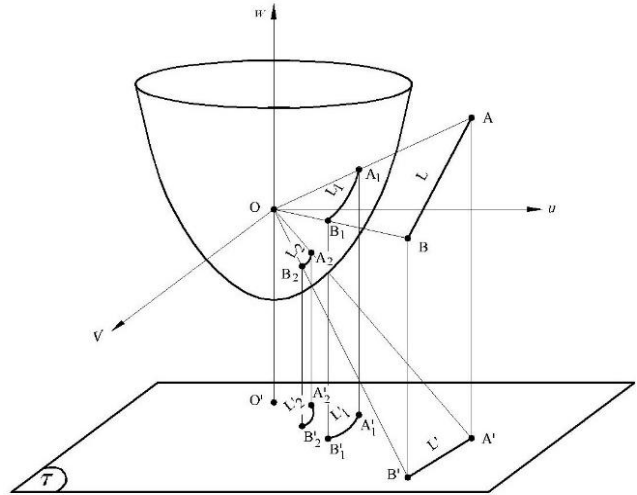

**Figure. 2. Projected lines onto paraboloid surface and directory plane.**

The desired curves to be studied are the planar curves on the paraboloid surface which are the perspective projection of space lines into the surface. Such curves may be circles, ellipses or parabolas according to the orientation of plane $\alpha$.

Plane α may be expressed as:

$\ell.U + m.V + n.W = 0$  Eq. (14).

where $\ell$, $m$, $n$ are the direction cosines of the normal to the plane and in the directions of $u$, $v$ and $w$ axis respectively.

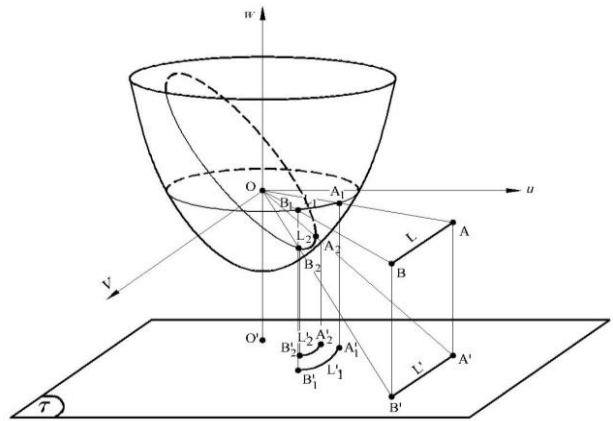

**Figure.3. Circle as a perspective of horizontal line.**

Figure .3 displays the circular curve of the intersection between the plane $\alpha$ [$O, L$] and the surface which is resulted when $n = \pm 1$ and both $\ell$ and $m = 0$, i.e., $L$ and $\alpha$ are horizontal. Hence: $L=L'$ which is given by Eq. (13), and

$L_1 = L_1' = 2f \cdot \Phi$  Eq. (15).

$\Phi = |\Phi_B - \Phi_A|$  Eq. (16).

where $\Phi$ is the central angle of the arc $L_1$:

$\Phi_A = \tan^{-1}\left(\dfrac{V_{A_1}}{U_{A_1}}\right)$  Eq. (17).

$\Phi_B = \tan^{-1}\left(\dfrac{V_{B_1}}{U_{B_1}}\right)$  Eq. (18).

While, $L_2$ is an elliptic arc and $L'_2$ is a circular arc, their lengths are to be introduced in the following case:

The elliptic arc of intersection is generated when:

-1 < $n$ < 1. Such curve may be extracted by modeling the curve as intersection between the surface and a right circular cylinder whose axis is parallel to *w*-axis and passes through point $O_E$, [1], where, $O_E$ is the Ellipse's center and its coordinates are:

$$U_{O_E} = -2f \frac{\ell}{n}$$ Eq. (19).

$$V_{O_E} = -2f \frac{m}{n}$$ Eq. (20).

$$W_{O_E} = 2f \frac{\ell^2 + m^2}{n^2}$$ Eq. (21).

And the radius of the cylinder is the distance $O'_E A'_1$.

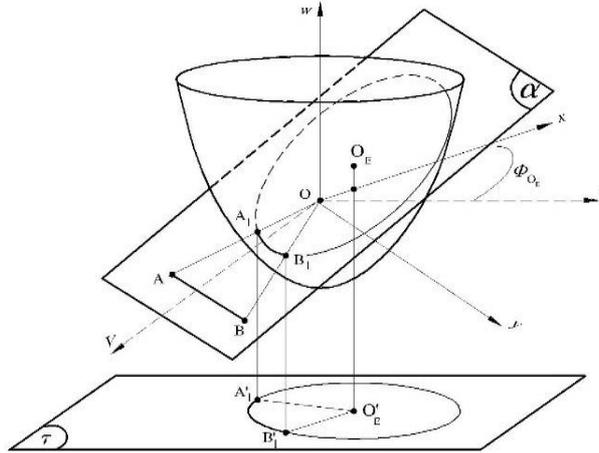

**Figure.4. Ellipse of intersection between paraboloid and plane.**

Since elliptic arc length is mainly depends on the major and minor axes lengths, then, such axes may be determined by rotating *u* and *v* axes by the angle $\Phi_{O_E}$ around *w*-axis, Figure.4. to direct the *x-z* plane in front view and viewing *α* as a line, i.e., edge view, Figure. 5. Where:

$$\Phi_{O_E} = \tan^{-1}\left(\frac{\ell}{m}\right)$$ Eq. (22).

According to [6], The resulted rotation matrix is:

$$\begin{bmatrix} X \\ Y \\ Z \end{bmatrix} = \begin{bmatrix} Cos(\Phi_{O_E}) & Sin(\Phi_{O_E}) & 0 \\ -Sin(\Phi_{O_E}) & Cos(\Phi_{O_E}) & 0 \\ 0 & 0 & 1 \end{bmatrix} \begin{bmatrix} U \\ V \\ W \end{bmatrix}$$ Eq. (23).

In Figure.5., the major axis length $2a_E$ is the distance between the two endpoints of the edge view $I_1$ and $I_2$ where:

$$2a_E = I_1 I_2 = 4f\left(1 + \frac{w_{O_E}^2}{r_{O_E}^2}\right)$$ Eq. (24).

And

$$r_{O_E} = \sqrt{U_{O_E}^2 + V_{O_E}^2}$$ Eq. (25).

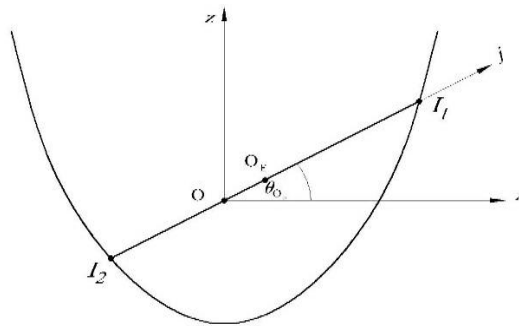

**Figure. 5. Ellipse's major axis in the x-z plane.**

Then, the minor axis may be determined as the line of intersection between plane α and plane α*, where α* is passing through OE and perpendicular to x-axis. Hence, α* is parallel to y-z plane, i.e., the minor axis is horizontal line parallel to y-z plane, Figure.6. the minor axis length may be expressed as:

$$2b_E = I_3 I_4 = 4f \sqrt{\left(1 + \frac{w_{O_E}^2}{r_{O_E}^2}\right)} \qquad \text{Eq. (26).}$$

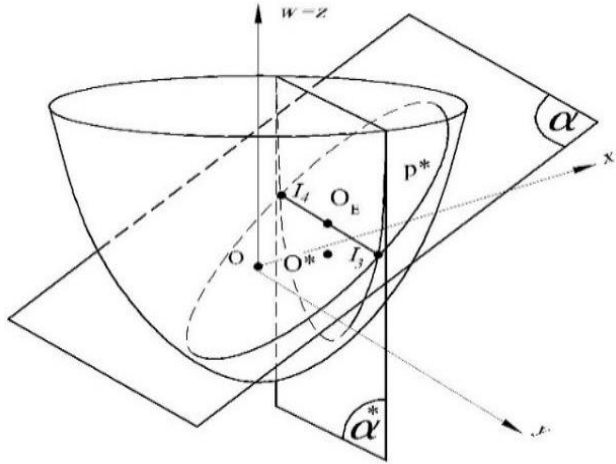

**Figure. 6. Minor axis length determination on parabola p*.**

A new local coordinate system ($O_E$, $j$, $k$) is to be originated in the plane $α$ considering point $O_E$, as the system origin, $j$-axis is directed towards the ellipse major axis and $k$-axis direction is perpendicular to $j$-axis.

To obtain the local coordinates ($J_{A_1}$, $K_{A_1}$) which is related to the given space point A ($U_A$, $V_A$, $W_A$), firstly we obtain the perspective projection $A_1$ of A then transform its coordinates into the local coordinate system as follows:

$$X_{A_1} = U_{A_1} \cos \Phi_{O_E} - V_{A_1} \sin \Phi_{O_E} \qquad \text{Eq. (27).}$$
$$Y_{A_1} = U_{A_1} \sin \Phi_{O_E} + V_{A_1} \cos \Phi_{O_E} \qquad \text{Eq. (28).}$$
$$Z_{A_1} = W_{A_1} \qquad \text{Eq. (29).}$$

Then, according to Figure.7,

$$J_{A_1} = g_{A_1} + g_{O_E} \qquad \text{Eq. (30).}$$

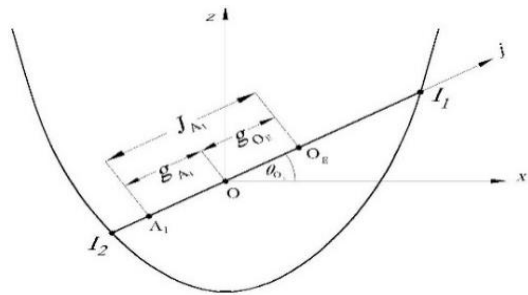

**Figure. 7. Local plane coordinates**

where:

$$g_{A_1} = \sqrt{X_{A_1}^2 + Z_{A_1}^2} \qquad \text{Eq. (31).}$$

$$g_{O_E} = \sqrt{X_{O_E}^2 + Z_{O_E}^2} \qquad \text{Eq. (32).}$$

The equation of the ellipse of intersection in the plane $α$ may be written as:

$$\frac{J_{A_1}^2}{a_E^2} + \frac{K_{A_1}^2}{b_E^2} = 1 \qquad \text{Eq. (33).}$$

From which:

$$K_{A_1} = \pm b_E \sqrt{1 - \frac{J_{A_1}^2}{a_E^2}}$$

Eq. (34).

$K_{A_1}$ will take the positive sign, since $A_1$ falls on the first Quadrant.

The elliptic arc length $L_1$ approximately equals to, [7]

$$L_1 = \left( \frac{\sqrt{(J_{B_1} - J_{A_1})^2 + (K_{B_1} - K_{A_1})^2}}{2 \sin\left(\frac{t_1 - t_2}{2}\right)} \right)(t_1 - t_2)$$

Eq. (35).

where:

$J_{A_1} = a_E \cdot \cos t_1$   Eq. (36).
$K_{A_1} = b_E \cdot \sin t_1$   Eq. (37).
$J_{B_1} = a_E \cdot \cos t_2$   Eq. (38).
$K_{B_1} = b_E \cdot \sin t_2$   Eq. (39).

$t_1$ and $t_2$ are two parameters.

Similarly, the true length of $L_2$, the perspective of the horizontal projection $L'$ on the directory plane is determined using Eq. (35) and the $(J, K)$ coordinates of $A_2$ and $B_2$.

The projected length $L'_1$ may be determined using the definite integration of the parabola's curve as follows:

$$L'_1 = r * \left| \sin^{-1}\left(\frac{U - U_{O'_E}}{r}\right) \right|_{U_{B'_1}}^{U_{A'_1}}$$

Eq. (40).

Where: $r$ is the radius of the circular arc $L'_1$.

The true length of $L_2$ and $L'_2$, Figure.2., may be determined similarly as the pervious analysis followed in determining $L_1$ and $L'_1$.

Finally, as $n = 0$, then a vertical plane $\alpha$ is resulted. In such case, parabolic curve of intersection $p$ is produced as shown in Figure.8. Alike the technique adopted in the preceding case, plane $\alpha$ is rotated by an angle $\Phi$, where $\Phi$ is the angle between $\alpha$ and the $u$-$w$ plane, then the true shape of the parabola could be viewed, Figure.9. Then, the coordinates of a points $A_1$ and $B_1$ on the curve may be rewritten, according to local coordinate system $\{O, x, z\}$, by substituting in Eqs. (27) and (29) with $U, V$ coordinates of $A_1$, $B_1$ and $\Phi$ as the angle of rotation. Hence, the horizontally projected length $L'_1$ between the two points $A'_1$ and $B'_1$ is:

$$L'_1 = |X_{A_1} - X_{B_1}|$$

Eq. (41).

and the true length of the curve $L_1$ along the parabola between $A_1$ and $B_1$ is:

$$L_1 = f\left[\sinh^{-1}\left(\frac{X}{2f}\right) + \sqrt{X^2 + 4f^2}\right]_{X_{B_1}}^{X_{A_1}}$$

Eq. (42).

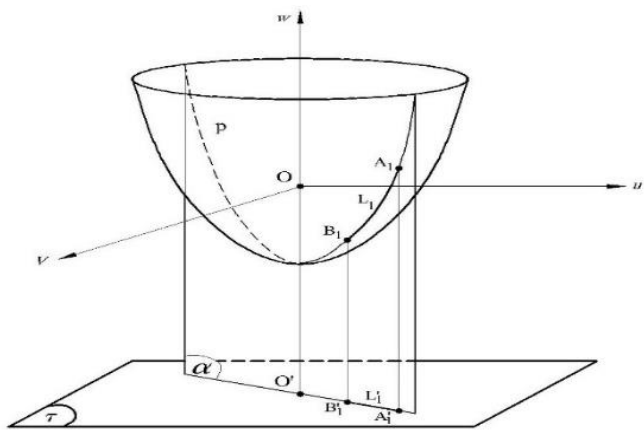  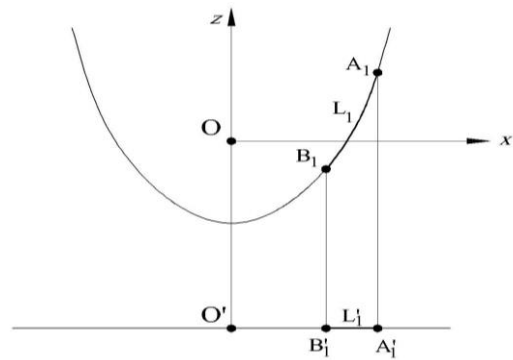

**Figure.8. Vertical plane intersection with paraboloid**    **Figure.9. True shape of the curve of intersection**

### *2.1.2) Area Involving Paraboloid Double Projection*

Another fundamental aspect characterizes metric properties of the paraboloid double projection is the area determination. Alike true length analysis, area determination incorporates determining area in space, perspectively projected area on the paraboloid surface and orthogonally projected area on the directory plane *τ* are to be manipulated. Two cases involving area determination will be studied herein. Figure.10. exposes a space area *ABCD* where the lines *AB* and *CD* are exhibits circular arcs whose center belongs to the surface axis, and the two lines *AD* and *BC* are vertical. Hence, such area is considered as a portion of vertical circular cylinder. The surface

$$a = (W_A − W_D). r_A . \lambda \qquad \text{Eq. (43).}$$

area of *ABCD* is:

where,
$r_A$ is given by Eq. (5), and:
$$\lambda = \frac{\pi}{180}.(\Phi_A − \Phi_B) \qquad \text{Eq. (44).}$$
Where,
$$\Phi_A = tan^{-1}(\frac{V_A}{U_A}) \qquad \text{Eq. (45).}$$
$$\Phi_B = tan^{-1}(\frac{V_B}{U_B}) \qquad \text{Eq. (46).}$$

Eventually, according to Figure.10, the perspective projection of the circular arcs *AB* and *CD* are circular arcs $A_1B_1$ and $C_1D_1$ [8], and perspective of vertical lines *BC* and *AD* are parabolic arcs $B_1C_1$ and $A_1D_1$. Area of the orthogonal projection of the region $A_1B_1C_1D_1$, i.e. the region $A'_1B'_1C'_1D'_1$, is denoted by χ, where:

$$\chi = \frac{\pi}{360}.(r_A^2 − r_C^2)(\Phi_A − \Phi_B) \qquad \text{Eq. (47).}$$

and, $r_C$ may be obtained by substituting with *C* coordinates in Eq. (5). Area of the perspective projection of the portion *ABCD*, i., e. area of the paraboloid portion $A_1B_1C_1D_1$, may be obtained employing the finite element technique as follows [6].

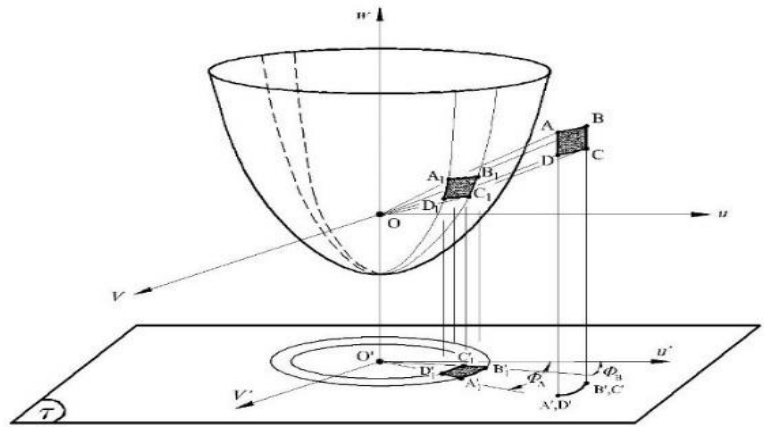

**Figure. 10. Area of cylindrical portion.**

- The orthogonal projection $A'_1B'_1C'_1D'_1$ is divided into mesh elements as shown in Figure. 11. Each element has dimensions of *Δr* and *ΔΦ*. [9], which are finite increments of radius *r* and angle *Φ* respectively, and:

$$\Delta r = \frac{r_A − r_C}{\text{Rows Number} − 1} \qquad \text{Eq. (48).}$$

$$\Delta \Phi = \frac{\Phi_A − \Phi_B}{\text{Colunmns Number} − 1} \qquad \text{Eq. (49).}$$

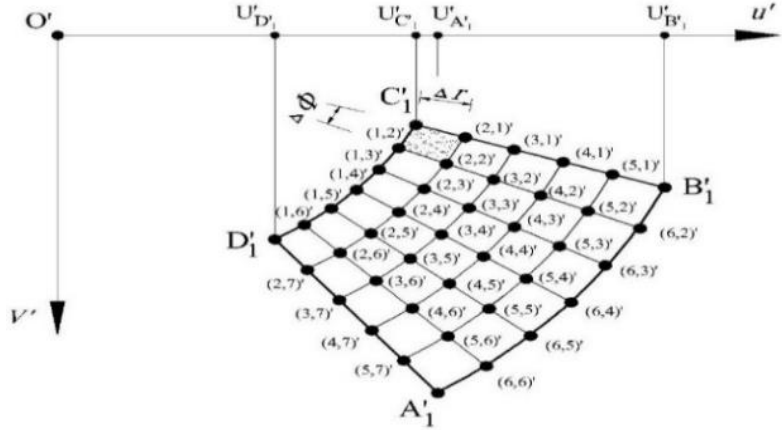

**Figure.11. Finite element meshing**

- Each unit area has four nodes, e.g. unit area *d(a′₁)* of nodes (*C′₁*), (*1,2*)', (*2,2*)' (*2,1*)' shown in Figure.12., each of which has its counterpart on the surface as the intersection of the surface and the perpendicular line to the plane $\tau$ from the node.

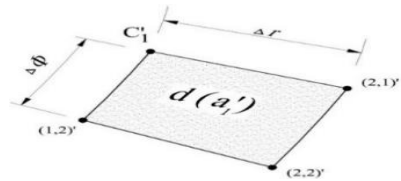

**Figure.12. Meshing of unit area**

-Then, coordinates of the nodes of the mesh unit region on the surface are expressed in polar coordinate form as:

$U_{C_1} = r_C \cos \Phi_B$ \hfill Eq. (50).

$V_{C_1} = r_C \sin \Phi_B$ \hfill Eq. (51).

and, substituting in the equation of the surface, Eq. (1),

$W_{C_1} = \frac{1}{4f}(r_C^2 - 4f^2)$ \hfill Eq. (52).

And for the node (*1,2*)

$U_{(1,2)} = r_C \cos(\Phi_B + \Delta\Phi)$ \hfill Eq. (53).

$V_{(1,2)} = r_C \sin(\Phi_B + \Delta\Phi)$ \hfill Eq. (54).

$W_{(1,2)} = \frac{1}{4f}(r_C^2 - 4f^2)$ \hfill Eq. (55).

and so, for the whole nodes of the mesh

- Area of the unit (*C₁*) (*1,2*) (*2,2*) (*2,1*) on the surface may be expressed as:

$d(a_1) = \frac{1}{2}\left(\overrightarrow{(2,2)(2,1)} \times \overrightarrow{(2,2)(1,2)} + \overrightarrow{(C_1)(2,1)} \times \overrightarrow{(C_1)(1,2)}\right)$ \hfill Eq. (56).

Where the vectors in the right-hand side of Eq. (64) are:

$\overrightarrow{(2,2)(2,1)} = (U_{(2,1)} - U_{(2,2)}, V_{(2,1)} - V_{(2,2)}, W_{(2,1)} - W_{(2,2)})$

$\overrightarrow{(2,2)(1,2)} = (U_{(1,2)} - U_{(2,2)}, V_{(1,2)} - V_{(2,2)}, W_{(1,2)} - W_{(2,2)})$

And

$\overrightarrow{(C_1)(2,1)} = (U_{(2,1)} - U_{(C_1)}, V_{(2,1)} - V_{(C_1)}, W_{(2,1)} - W_{(C_1)})$

$\overrightarrow{(C_1)(1,2)} = (U_{(1,2)} - U_{(C_1)}, V_{(1,2)} - V_{(C_1)}, W_{(1,2)} - W_{(C_1)})$

Finally, the area of the region $A_1B_1C_1D_1$ is the summation of the area of the regions of Eq. (57), since:

$A_1B_1C_1D_1 = \sum_{a_i}^{a_1}(d(a_1) + d(a_2) + d(a_3) + \cdots + d(a_i))$ \hfill Eq. (57).

Where i is the total number of the meshes.

Another case study is determination the area of a vertical rectangle *ABCD* as shown in Figure. 13.

In the Figure, the perspective of the two vertical lines *AD* and *BC* are parts of parabolic curves $A_1D_1$ and $B_1C_1$, while those of horizontal sides *AB* and *CD* are parts of either circles or ellipses as previously indicated.

Orthogonal projection *A′₁B′₁C′₁D′₁* of the perspective

$A_1B_1C_1D_1$ is a region shown in Figure.14., which is bounded by the two lines $A'_1D'_1$ and $B'_1C'_1$, and the two circular arcs $A'_1B'_1$ and $C'_1D'_1$.

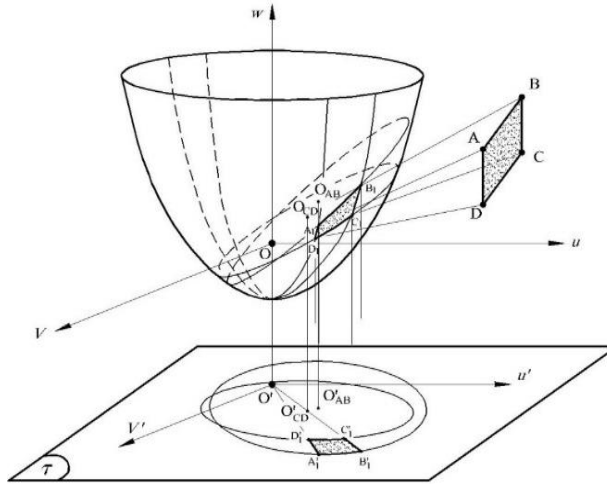

**Figure.13. Projection of vertical area**

The projected area is determined by dividing such area into three regions according to Figure.15.

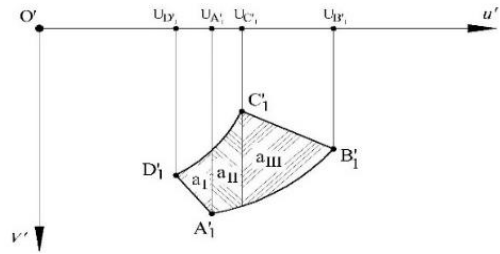

**Figure. 14. Orthogonally projected area of the perspective projection of vertical rectangle**.

- **Region I** which is bounded by the circular arc $C'_1D'_1$, vertical line $A'_1 U'_{A'_1}$, and the line $A'_1D'_1$.
- **Region II** which is bounded by the two circular arcs $C'_1D'_1$ and $A'_1B'_1$, and two vertical lines $A'_1 U'_{A'_1}$ and $C'_1 U'_{C'_1}$.
- **Region III** which is bounded by the circular arc $A'_1B'_1$, vertical line $C'_1 U'_{C'_1}$, and the line $C'_1B'_1$.

For *Region I*, equation of the line $A'_1D'_1$ is:

$$U + s^*.V = 0 \qquad \text{Eq. (58)}.$$

Where: $s^* = \dfrac{-U_{A'_1}}{V_{A'_1}}$

and the circular arc $C'_1D'_1$ equation is:

$$(U - U_{O_{CD}})^2 + (V - V_{O_{CD}})^2 = R_{CD}^2 \qquad \text{Eq. (59)}.$$

where $R_{CD}$ is the radius of the circular arc $C'_1D'_1$, Eq. (5), $U_{O_{CD}}$ and $V_{O_{CD}}$ are $u, v$ – coordinates of the circular arc center, Eqs. (19) and (20). Then, the area $a_I$ of the portion *I* may be expressed as:

$$a_I = \int_{U_{D'_1}}^{U_{A'_1}} -\left(s^*.V\left(\sqrt{R_{CD}^2 - (V - V_{O_{CD}})^2} + U_{O_{CD}}\right)\right) dV \qquad \text{Eq. (60)}.$$

For the *Region II*, equation of the arc $A'_1B'_1$ is:

$$(U - U_{O_{AB}})^2 + (V - V_{O_{AB}})^2 = R_{AB}^2 \qquad \text{Eq. (61)}.$$

where $R_{AB}$ is the radius of the circular arc $A'_1B'_1$, Eq. (5), $U_{O_{AB}}$ and $V_{O_{AB}}$ are $u, v$ – coordinates of the circular arc center, Eqs. (19) and (20).

Then, the area $a_{II}$ of portion *II* may be expressed as:

$$a_{II} = \int_{U_{A'_1}}^{U_{C'_1}} \left(-\left(\left(\sqrt{R_{AB}^2 - (V - V_{O_{AB}})^2} + U_{O_{AB}}\right) - \left(\sqrt{R_{CD}^2 - (V - V_{O_{CD}})^2} + U_{O_{CD}}\right)\right)\right) dV \qquad \text{Eq. (62)}.$$

For the third *Region III*, the equation of the line $B'_1C'_1$ is:

$U + s^{**}.V = 0$  Eq. (63).

where: $s^{**} = \dfrac{-U_{B'_1}}{V_{B'_1}}$

Then, the area $a_{III}$ of the portion *III* may be expressed as:

$$a_{III} = \int_{U_{D'_1}}^{U_{A'_1}} \left( \left( \sqrt{R_{AB}^2 - (V - V_{O_{AB}})^2} + U_{O_{AB}} \right) + S^{**}.V \right) dV \qquad \text{Eq. (64).}$$

Hence, $a_{I,\,II,\,III}$, the orthogonally projected region area $A'_1B'_1C'_1D'_1$ is:

$a_{I,II,III} = a_I + a_{II} + a_{III}$  Eq. (65).

Alike the presented first example, the finite element technique is traced to determine the perspectively projected area $A_1B_1C_1D_1$ on the paraboloid surface.

In favor of generating the mesh Figure. 14. shows that the two circular arcs $A'_1B'_1$ and $C'_1D'_1$ in the directory plane $\tau$ are not concentric.

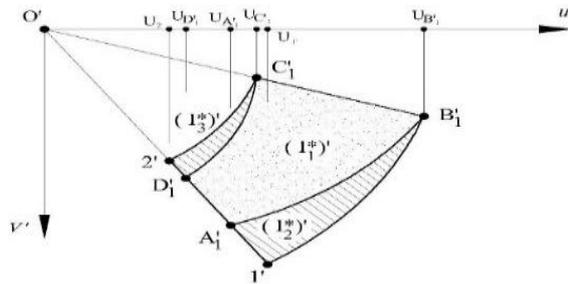

**Figure.15. Projected area components**

So, the mesh would be easily generated by dividing the whole region into three portions $I_1^{*'}$, $I_2^{*'}$, and $I_3^{*'}$, then the desired area $I_1^*$ is the subtraction of the areas $I_2^*$ and $I_3^*$ from the total area.

The whole area $a_{Total}$, the area of region $1B_1C_12$, on the paraboloid surface matches the area $A_1B_1C_1D_1$ in Figure. 10. whose mesh generation pattern is indicated in Figure. 11., and its value is given by Eq. (57).

Area $I_2^*$ and $I_3^*$ on the paraboloid surface may be determined by constructing a mesh for their corresponding projected area on the directory plane $\tau$, i.e. $I_2^{*'}$, and $I_3^{*'}$. The finite element net for the region $I_2^{*'}$ is constructed by dividing the central angle $B'_1 O'_{AB} A'_1$ into finite angles to get the net shown in Figure. 16., then, the area of each net division could be easily obtained using the technique tracked in Eq. (56). and Eq. (57). to achieve the area $I_2^*$ on the paraboloid surface. The same analysis is tracked to obtain the area of the region $I_3^*$ on the surface, whose corresponding projected area on the directory plane $\tau$ is the region $2'C'_1D'_1$. Hence, the desired area $I_1^*$ of the perspective projection of the vertical rectangle *ABCD* is:

$a_{I_1^*} = a_{Total} - \left( a_{I_2^*} + a_{I_3^*} \right)$  Eq. (66).

Area of the perspective of the projected area may be determined in similar way as previously analyzed.

As a conclusion, the finite element technique may be handled to obtain the area of the perspective projection of any plane figure whose sides are straight lines, whether it is in general position or in particular, as horizontal for example, as well as its orthogonal projection area on the directory plane $\tau$.

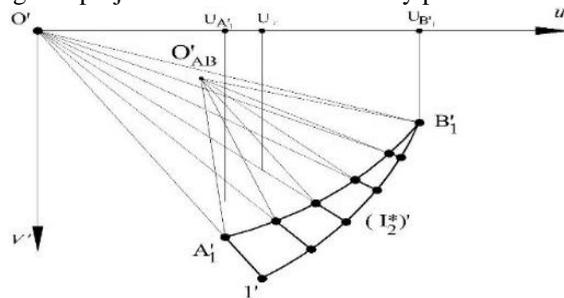

**Figure.16. $I_2^{*'}$ division**

## 3) CONCLUSIONS

Parabolic double projection may be considered as the most relevant projection system for the catadioptric camera imaging in the favor of wide viewing coverage. Analytical analysis of such projection involves orthographic and perspective projections onto the directory plane and on the surface respectively employing the analytical analysis for determining the projection metric properties. Such presented analysis is genuine and is considered as an essential tool for camera calibration. Metric properties analyzed herein are the determination of the space and both orthographically and perspectively projected true lengths and areas. Perspective projection of lines on paraboloid surface may be circles, ellipses or parabolas according to their plane orientation, hence their true length is determined analytically in direct manner. Three different lines are presented for line true length determination. Since areas, either those projected on the surface or on the directory plane, are bounded by conic curves. Hence, such areas may be determine using the finite element technique. Different plane and one cylindrical area were analyzed utilizing such technique.

## *4) Refrences*